%% file: acl2023.tex
\pdfoutput=1
\documentclass[11pt]{article}
\usepackage[]{ACL2023}
\usepackage{times}
\usepackage{latexsym}
\usepackage{xspace}
\usepackage[T1]{fontenc}
\usepackage[utf8]{inputenc}
\usepackage{microtype}
\usepackage{inconsolata}
\usepackage{amssymb}
\usepackage{amsmath}
\usepackage{multirow} 
\usepackage{graphicx} 
\usepackage{amsthm}
\usepackage{bbm}
\theoremstyle{definition}
\newtheorem{definition}{Definition}[section]
\newtheorem{theorem}{Theorem}
\newtheorem{lemma}[theorem]{Lemma}
\newcommand\framework{\texttt{DivDist}\xspace}
\newcommand\soa{\texttt{SoA}\xspace}
\newcommand\normalize{\texttt{normalize}\xspace}
\newcommand\divergence{\text{$D$}\xspace}

\def\bias{\qopname\relax m{\texttt{bias}}}

\usepackage{bbm}
\usepackage{amsfonts}
\usepackage{booktabs}
\usepackage{url}
\usepackage{nicefrac}

\def\Snospace~{\S{}}

\newcommand\eg{e.g.\xspace}
\newcommand\ie{i.e.\xspace}
\newcommand\cf{cf.\xspace}

\input std-macros
\input macros

\title{Trustworthy Social Bias Measurement}

\author{
Rishi Bommasani \\
Stanford University \\
\texttt{nlprishi@stanford.edu} \\
\And
Percy Liang \\
Stanford University \\
\texttt{pliang@cs.stanford.edu} \\}

\begin{document}
\abovedisplayskip=1mm
\abovedisplayshortskip=2.0mm
\belowdisplayskip=1mm
\belowdisplayshortskip=2.0mm
\maketitle

\input{sections/0-abstract}
\input{sections/1-introduction}
\input{sections/2-principles}
\input{sections/3-measurement-framework}
\input{sections/4-bias-measures}
\input{sections/4-measurement-verification}
\input{sections/5-validity}
\input{sections/6-reliability}
\input{sections/7-related-work}
\input{sections/8-measure-modeling}
\input{sections/9-conclusion}
\input{sections/10-reproducibility}

\section*{Acknowledgements}
\input{sections/acknowledgements.tex}

\bibliography{all, anthology}
\bibliographystyle{acl_natbib}

\appendix
\input{appendices/01_reproducibility.tex}

\end{document}

%% file: std-macros.tex
\ifthenelse{\isundefined{\definition}}{\newtheorem{definition}{Definition}}{}
\ifthenelse{\isundefined{\assumption}}{\newtheorem{assumption}{Assumption}}{}
\ifthenelse{\isundefined{\hypothesis}}{\newtheorem{hypothesis}{Hypothesis}}{}
\ifthenelse{\isundefined{\proposition}}{\newtheorem{proposition}{Proposition}}{}
\ifthenelse{\isundefined{\theorem}}{\newtheorem{theorem}{Theorem}}{}
\ifthenelse{\isundefined{\lemma}}{\newtheorem{lemma}{Lemma}}{}
\ifthenelse{\isundefined{\corollary}}{\newtheorem{corollary}{Corollary}}{}
\ifthenelse{\isundefined{\alg}}{\newtheorem{alg}{Algorithm}}{}
\ifthenelse{\isundefined{\example}}{\newtheorem{example}{Example}}{}
\newcommand{\E}{\ensuremath{\mathbb{E}}} 

%% file: macros.tex
\definecolor{mina-blue}{rgb}{0.3333, 0.3764, 0.6627}
\definecolor{mina-green}{rgb}{0.4078, 0.6392, 0.4274}
\definecolor{mina-yellow}{rgb}{0.9803, 0.7372, 0.0352}
\definecolor{mina-orange}{rgb}{0.9450, 0.5647, 0.1254}

\newcommand{\best}[1]{\colorbox{mina-green!30}{#1}}
\newcommand{\worst}[1]{\colorbox{pink!50}{#1}}

%% file: sections/0-abstract.tex
\begin{abstract}
\label{abstract}
How do we design measures of social bias that we \textit{trust}?
While prior work has introduced several measures, no measure has gained widespread trust: instead, mounting evidence argues we should distrust these measures.
In this work, we design bias measures that warrant trust based on the cross-disciplinary theory of measurement modeling.
To combat the frequently fuzzy treatment of social bias in NLP, we explicitly define social bias, grounded in principles drawn from social science research.
We operationalize our definition by proposing a general bias measurement framework \framework, which we use to instantiate  5 concrete bias measures.
To validate our measures, we propose a rigorous \textit{testing protocol} with 8 testing criteria (\eg predictive validity: do measures predict biases in US employment?).
Through our testing, we demonstrate considerable evidence to trust our measures, showing they overcome conceptual, technical, and empirical deficiencies present in prior measures.
\end{abstract}

%% file: sections/1-introduction.tex
\section{Introduction}
\label{sec:introduction}
Language technologies are increasingly critical to our lives and to broader societal function.
As NLP researchers, our work has increasingly direct, immediate, and significant impact: we must reckon with this and, especially, any harms that arise from language technology.
Social bias is a central consideration \citep[][\textit{inter alia}]{hovy2016, bender2021, weidinger2022}: how we represent people and what we associate them with has material consequences.
Biased language technology can cause several types of harm \citep[][\S5.1]{dev-etal-2022-measures, bommasani2021}:
allocational (\eg~lower hiring rates for marginalized groups due to algorithmic resume screening), 
representational (\eg~associating Muslims with violence), 
and psychological (\eg~stereotype threat).

\textit{Measurement} functions as the primary lens for understanding social bias in NLP.
And measurement is seen as an essential to successfully reducing bias: to determine if an intervention mitigates bias, the measured bias should decrease due to the intervention.
If all paths forward for making progress on bias in NLP pass through measurement, then what is the current state of bias measurement?

Many works have proposed bias measures, spanning different settings like text, vector representations, language models, and task-specific models \citep[see][]{blodgett2020, dev-etal-2022-measures}.
Most measure bias between two social groups.
However, no standard exists for what evidence is required to trust these measures: works provide a mixture of intuitive, empirical, and theoretical justifications.
Perhaps as a consequence, many works are subject to scrutiny: measures have been shown to be brittle \citep{ethayarajh2019, nissim2020, antoniak2021, delobelle-etal-2022-measuring}, contradictory \citep{bommasani2020}, unreliable \citep{aribandi-etal-2021-reliable, seshadri2022}, invalid \citep{blodgett2021}, and the space overall is unclear on what bias means and what metrics purport to measure \citep{blodgett2020}. 
Trust is necessary: for metrics to productively guide progress and inform decision-making, we must trust them.

Consequently, we focus on \textit{trustworthy} bias measurement.
We apply \textit{measurement modeling} to address this challenge: measurement modeling is an expansive theory used across the social sciences to design and validate measures of (complex) social constructs.
Therefore, measurement modeling is well-suited to social bias measurement: the theory has a longstanding tradition for similar social constructs, including even for social bias in humans \citep[\eg the Implicit Association Test;][]{chequer2014}. 

Under measurement modeling, we must first define the \textit{theoretical construct} of social bias.
In contrast, \citet{blodgett2020} showed many works in NLP failed to (adequately) define social bias.
To define social bias, we draw upon principles in social science research; these principles dictate how we operationalize our definition into a \textit{general} measurement framework.
Our measurement framework \framework, based on divergences between probability distributions, improves over prior work in two key ways: (i) \textit{compatibility}, meaning bias measures can be instantiated for several settings (\eg text, vector representations) that yield comparable measurements and (ii) \textit{multi-group}, meaning we are not restricted to binary bias measurement. 
These properties are valuable for NLP: for example, we may want to understand how different processes change biases (\eg the potential bias amplification between training data and a learned model, or between a generative model and its samples).

Beyond offering generality, our framework also makes explicit that bias is fundamentally a \textit{relative} phenomenon, which has been neglected in all prior work. 
To meaningfully measure bias, one must state the \textit{normative} reference frame: what would constitute (no) bias?
This is a material consideration: the relevant reference could be a particular ideal (\eg equal association across groups), social status quo (\eg the US labor demographics), or technical contrast (\eg a model's training data), but regardless the choice determines what bias even means. 
By allowing the reference to be specified, rather than being assumed, our framework enables pluralism: different normative positions can dictate what constitutes bias.

Using our framework, we instantiate 5 bias measures, spanning measures for text, word embeddings, and contextualized representations.
We put these measures to the test, alongside several prior measures.
Measurement modeling specifies 8 well-studied desiderata: in a sense, measurement modeling provides a well-established checklist of criteria to trust measures of social constructs.
For each desideratum, we design a test, amounting to the first rigorous \textit{testing protocol} for validating bias measures.
Executing these tests, we accrue evidence to trust our measures, while surfacing concerns with prior measures. 

Beyond our primary contributions (measurement framework, testing protocol), we make several striking findings while testing our measures.
First, our bias measure for word embeddings strongly correlates with societal trends in employment, whereas some prior measures are uncorrelated or even \textit{anti-correlated}, suggesting our measure is more appropriate for certain computational social science applications.
Second, our measures indicate the representations in GPT-2 \citep{radford2019} amplify biases relative to GPT-2's training data, but this amplification remains latent and unobserved when sampling from the model, which poses broader questions regarding how biases acquired in training language models propagate downstream \citep{goldfarb2021, steed2022}.
Third, "debiasing" methods generally fail to reduce, and sometimes \textit{exacerbate}, social bias according to our measure, which calls into question their meaningfulness \citep{gonen2019}.

%% file: sections/2-principles.tex
\section{Principles for Social Bias}
\label{sec:principles}
\noindent \textbf{Notation.}
Following social science research, we define social bias in terms of social \textit{groups} ${G}_1, \dots, {G}_k$,\footnote{We acknowledge that many categories (\eg race, gender) are the subject of abundant disagreement \citep{crenshaw1989, penner2015}.} which reflect a categorization of individuals \citep{allport1954}, and a \textit{target concept} ${T}$, which bias is measured with respect to.
As an example, we may consider the gender biases in science with $G_1 = \text{female}, G_2 = \text{male}$ and $T = \text{scientist}$.  

\noindent \textbf{Reducing bias to associations.}
Given social groups and a target concept, some social science theories define bias as the target concept's differential \textit{association} with each group.
For example, in the Implicit Association Test \citep{greenwald1998}, the test uses response time to quantify the association between the target concept and each group.
Further, these associations must be \textit{systematic}: 
\citet{beukeboom2019} write that "bias is a \textit{systematic} asymmetry", and \citet{friedman1996} emphasize that social bias pertains to broader social groups, rather than particular individuals \citep[\cf][]{bommasani2022picking}. 

\subsection{Bias is Relative}
\label{subsec:bias-attribution}
If a machine translation model exactly replicates the properties of its training data, is it biased?
It depends.
Relative to its training data, no, but relative to a specific societal reference, potentially yes, namely if the training data was biased with respect to this reference.
Most bias measures in NLP ignore this fundamental property: bias is, instead, portrayed as absolute by many measures.

This fundamentally misconstrues what bias is: bias is an inherently \textit{relative} construct, which requires that a \textit{reference} be specified.
Bias is precisely the extent to which the observed associations diverge from this reference. 
Since bias emerges through social processes, reference-sensitive measures allow us to understand how different decisions increase/reduce bias \citep{friedman1996}. 
In this spirit, \citet{shah2020} and \citet{hovy2021} attribute bias in NLP to several sources (\eg data selection, data annotation, model training): effective bias measurement could hope to quantify the relative contribution of each of these sources. 

\subsection{Defining Social Bias}
\label{subsec:bias-definition}
Having introduced groups, targets, associations, and references, we define social bias.
\begin{definition}[Social Bias]
\label{defn:relative-social-bias}
\textit{Social bias} is the divergence in the observed associations between a target concept and a set of social groups from corresponding reference associations.
\end{definition}
In particular, most works in NLP and the social sciences construe social bias as an "asymmetry" in the observed associations (e.g. the bias that \textit{scientist} is more associated with the male gender than the female gender), as in \citet{beukeboom2019}.
This perspective on bias is a special case of our definition, when the reference is the uniform baseline: no bias corresponds to the target concept being equally associated with every social group.

%% file: sections/3-measurement-framework.tex
\section{\framework Measurement Framework}
\label{subsec:our-framework}
Having defined social bias, we propose our two-stage measurement framework \framework.
First, given \textit{parameters}, which specify the associations of interest, \framework yields a bias measure $\bias$.
Second, given \textit{inputs} (\ie the target concept, social groups, and reference mentioned in our distribution), $\bias$ yields a bias measurement $\bias(T, G_1, \dots, G_k; \mathbf{p}_0)$ (\ie a numerical value of how much bias is present).
\begin{align}
    \mathbf{s} \triangleq~&[\soa(T, G_1), \dots, \soa(T, G_k)]  \\
    \mathbf{p} \triangleq~& \normalize(\mathbf{s}) \\
    \bias&(T, G_1, \dots, G_k; \mathbf{p}_0) = \divergence(\mathbf{p}, \mathbf{p}_0)
\end{align}
\noindent \textbf{Parameters.}
To map from the abstract framework \framework to a concrete bias measure $\bias$, we specify three functions (\soa, \normalize, \divergence).
First, \soa quantifies the strength of association between the target concept and a social group as a numerical value in $\mathbb{R}_{\geq 0}$ 
This function handles both setting-specific aspects of measurement (\ie \soa is considerably different for text vs. vector representations) and the specific associations of interest (\eg different \soa implementations are needed to measure frequency-related biases vs. more semantic biases).
Applying \soa to every (target concept, social group) pair yields the observed association vector $\mathbf{s} \in \mathbb{R}_{\geq 0}^k$.
Second, we normalize $\mathbf{s}$ to a categorical distribution $\mathbf{p}$ using \normalize.
Third, we quantify the divergence using \divergence between the (normalized) observed associations $\mathbf{p}$ and the reference associations $\mathbf{p}_0$, which we also specify as a categorical distribution distributed over the groups. 

Observe the clear correspondence between our framework \framework and our definition: 
Step 1 extracts the observed associations, 
Step 2 prepares these associations, and 
Step 3 measures the divergence from reference associations. 
This correspondence indicates our measures demonstrate \textit{structural fidelity} \citep{loevinger1957}, one of 8 desiderata we consider in measurement modeling.

\noindent \textbf{Inputs.}
To further map from the bias measure $\bias$ to the bias measurement $\bias(T, G_1, \dots, G_k; \mathbf{p}_0)$, we specify three inputs $(T, G_1, \dots, G_k; \mathbf{p}_0)$.
In many cases, we will represent social groups ${G}_1, \dots, {G}_k$ and the target concept ${T}$ using \textit{word lists}, \ie representative words that embody the associated concept.
Further, for the reference $\mathbf{p}_0$, we will specify it as a categorical probability distribution distributed over the $k$ social groups, which encodes the association between each group and the target concept when there is no bias.


\noindent \textbf{Generality.}
We prove several prior bias measures, across the social sciences \citep[\eg][]{weitzman1972, voigt2017} and NLP \citep[\eg][]{caliskan2017, garg2018} are special cases of \framework. 
In all of these works, bias is measured in the binary setting as the difference in associations (\ie how much is the \textit{male} gender associated with \textit{scientist} more than the \textit{female} gender is associated with \textit{scientist}).
We show this interpretation of bias as a "systematic asymmetry" \citep{beukeboom2019} is recovered by \framework using the uniform distribution $\mathbf{p}_0 = \left[\frac{1}{2}, \frac{1}{2}\right]$, up to scaling.\footnote{For brevity, we abbreviate $\soa(G_1, T)$ and $\soa(G_2, T)$ as $x$ and $y$, respectively. WLOG, let $x \geq y$.}
\begin{align*}
\bias_{\text{prev}} &= \soa(T, G_1)~-~\soa(T, G_2)~=~x-y \\
\bias_{\text{ours}} &= D\left(\normalize\left(\left[x, y\right]\right), \left[\frac12, \frac12\right]\right) \\
&= \left\|\left[\frac{x}{x+y}, \frac{y}{x+y}\right] - \left[\frac{1}{2}, \frac{1}{2}\right]\right\|_1
\\
&= \frac{x-y}{x+y}
\end{align*}

%% file: sections/4-bias-measures.tex
\section{Measures}
\label{sec:measures}
To further demonstrate the generality of \framework, we instantiate several bias measures using it.
In NLP, we want to measure in a variety of settings: here, we introduce measures for (human-authored or machine-generated) text, (static) word embeddings, and contextualized representations to provide broad coverage. 
These measures differ in the implementation of the \soa parameter, which encodes the specifics of each setting; the choices for \normalize and \divergence are consistent across settings.

\subsection{Text}
\label{subsubsec:soa-txt}
Since social bias is a systematic phenomenon, bias in text manifests in distributional statistics.
To implement $\soa_{\text{text}}$, we quantify associations in text based on these statistics.
The association between concept $T$ and group $G_j$ in a corpus $\mathcal{L}$ is quantified as follows:
\begin{enumerate}
    \item Select contexts $c_1, \dots, c_N$ in corpus $\mathcal{L}$ such that each context $c_i$ mentions concept $T$. 
    \item For each context $c_i$,
    let $a_{ij} \in \{0,1\}$ indicate if $T$ is associated with $G_j$ in $c_i$.
    \item $\soa_{\text{text}}(T, G_j) = \sum_{i=1}^Na_{ij}$.
\end{enumerate}

\noindent \textbf{Contexts, mentions, and associations.}
The aforementioned procedure partially implements $\soa_{\text{text}}$, but leaves ambiguous: (i) what are contexts $c_i$, 
(ii) what does it mean for a concept $T$ to be mentioned in $c_i$, 
and (iii) what does it mean for group $G_j$ to be associated with $T$ in $c_i$?
For contexts, we consider three-sentence spans in $\mathcal{L}$ by default, testing the sensitivity of measurements to this choice subsequently.
For mentions, these judgments could ideally be made by human domain-experts, but since this is costly for large corpora, we automate this by requiring that we have a word list $W(T)$ for $T$ such that for each context $c_i, \exists w \in W(T)~s.t.~w \in c_i$.
For associations in a context, we consider two options.
In the \textbf{human} variant of our text bias measure, humans make these judgments, whereas in the \textbf{automated} variant, we require that some word in the group's word list $W(G_j)$ appears \textit{and} that no word in any other group's word list appears.\footnote{In pilot experiments measuring bias in English Wikipedia, the second constraint increased the precision of our heuristic (since contexts are more unambiguously associated with the group) with fairly marginal cost in recall.}

\subsection{Word Embeddings}
\label{subsubsec:soa-we}
For word embeddings, we quantify associations using cosine similarity, which is the standard similarity metric for word embeddings \citep[e.g.][]{mikolov2013} that has garnered some theoretical justification \citep{zhelezniak2019}.
Let $\mathbf{w}$ be the word vector for word $w$. 
We quantify the strength of association for word embeddings as
$$
    \soa_{\text{WE}}(T, G_j) = \cos\left(
    \E_{t \in W(T)}\mathbf{t}, \E_{g \in W(G_j)}\mathbf{g}
    \right).
$$
In words, $\soa_{\text{WE}}$ is the cosine similarity between the average target word embedding and average group word embedding, closely resembling  how \citet{caliskan2017} and \citet{garg2018} quantify association.

\subsection{Contextualized Representations}
\label{subsubsec:soa-cr}
For contextualized representations, most prior measures \citep[e.g.][]{may2019, tan2019, guo2020} compute a single bias value for these representations.
We argue this is a type error: the bias in contextualized representations will depend on the context in which the representations are used and, in fact, \citet{ethayarajh2019b} showed these representations are highly context-sensitive.
For example, the gender biases in BERT representations \citep{devlin2019} will differ when using BERT to embed text from the New York Times vs.~a misogynistic subreddit. 
With that in mind, we present two context-sensitive approaches that quantify strength of association in contextualized representations $\mathbf{w}_i$, which embed word $w$ in context $c_i$ within a text corpus $\mathcal{D}$. 

\noindent \textbf{Reduction to $\soa_{\text{WE}}.$}
Our first approach reduces the contextualized case to the static case, following \citet{bommasani2020}. 
For each (group or target) word $w$ of interest, we compute $\mathbf{w} = \E_{c_i \in \mathcal{D} \mid w \in c_i}\mathbf{w}_i$
as the average of $w$'s contextualized representations across all contexts in which it appears in corpus $\mathcal{D}$. 
\citet{bommasani2020} show this produces high-quality static embeddings from contextualized representations: once we have these static embeddings, we then apply the aforementioned $\soa_{\text{WE}}$ to quantify strength of association for contextualized representations. 

\noindent \textbf{Probing.}
The key downside to the reduction approach is the reduction may distort associations in the original contextualized representations \citep{bommasani2020}.
Therefore, we consider more direct techniques for interpreting contextualized representations \citep[see][]{rogers2020, belinkov2021}, which closely resemble the \textit{probing} \citep{alain2017, hewitt2019b} methodology studied in the interpretability community. 
Namely, we learn a classifier $f$ over the representations that simulates the human annotator from the text setting.

\noindent \textit{Training.}
$f$ receives a contextual vector $\mathbf{t}_i$ as input and predicts which social group (if any) the target word $t$ is associated with in context $c_i$. 
To assemble $f$'s training data, we 
(i) sample $N$ contexts $c_i$ that mention $T$ in the corpus $\mathcal{D}$, 
(ii) have humans annotate labels $y_i$ indicating which group $G_j$ (if any) that $T$ is associated with in $c_i$, 
(iii) embed the contexts $c_i$, 
and (iv) extract any contextual representations $\mathbf{t}_i$ for words $t \in W(T)$.
(Note that in step (ii), the human annotations can also be re-purposed to measure bias in the text corpus $\mathcal{D}$ itself.)
The resulting $\{(\mathbf{t}_i, y_i)\}_{i=1}^N$ examples are used to learn $f$ by minimizing the cross-entropy loss of predicting group labels from the corresponding contextualized representation.

\noindent \textit{Inference.}
To quantify strength of association, we sample further contexts $c^{\text{test}}$ that mention $T$:
\begin{align*}
    \soa_{\text{CR-Probe}}(T, G_j) = \sum_{i=1}^N
    \mathbbm{1}\left[{f(\mathbf{t_i})= G_j}\right].
\end{align*}
\noindent \textit{Decisions.}
Selecting the complexity of the classifier (i.e.~probe) have been the subject of intense debate in the probing community \citep{hewitt2019b, pimentel2020b, pimentel2020a, belinkov2021, hewitt2021, pimentel2021}.
We choose to learn linear classifiers, which indicates that we prioritize easily (i.e.~linearly) decoded associations \citep{ivanova2021, hewitt2021}. 
\input{tables/measures.table}

\subsection{Normalization and Divergence Parameters}
\label{subsec:additional-considerations}
Beyond \soa, \framework requires normalization $\normalize:~\mathbb{R}^k \to \Delta^{k-1}$ and divergence $\divergence:~\Delta^{k-1} \times \Delta^{k-1} \to \mathbb{R}_{\geq 0}$ to fully instantiate bias measures.
For conceptual simplicity, as defaults, we set \normalize to be dividing a vector by its sum (since the input vectors are generally/always non-negative, so this is a valid means for yielding a probability distribution) and \divergence to be the $\ell_1$ distance as a well-known and simple-to-understand divergence.
These settings correspond to our proof, where we show our measure generalizes several prior measures, and we show that our measurements are quite robust to these choices empirically in our sensitivity analysis (\autoref{sec:reliability}). 

%% file: tables/measures.table.tex
\begin{table*}[htp]
\resizebox{\textwidth}{!}{
\centering
\vspace{0mm}
\begin{tabular}{c|c|l}
\toprule
\multicolumn{1}{c}{\textbf{Setting}} & \multicolumn{1}{c}{\textbf{Abbreviation}} &
\multicolumn{1}{c}{\textbf{Implementation of \soa}}\\
\midrule
Text & Human & Number of contexts where $T$ and $G_j$ are associated based on human annotator \\
Text & Aut. & Number of contexts where $T$ and $G_j$ are associated based on cooccurrence \\
WE & Emb. & Cosine similarity between average embeddings for $W(T)$ and $W(G_j)$ \\
CR & Red. & Cosine similarity between representations averaged across contexts for $W(T)$ and $W(G_j)$ \\
CR & Probe & Number of contexts where $T$ and $G_j$ are associated based on learned probe \\
\bottomrule
\end{tabular}}
  \caption{Summary of the \textbf{implementations of \soa} we introduce in \autoref{sec:measures} for each setting.}
\label{tab:measures}
\end{table*}

%% file: sections/4-measurement-verification.tex
\section{Testing Protocol}
\label{sec:verification}
\input{tables/measurement-modeling.table}

We have put forth yet another bias measure(s); why should we trust it?
We turn to the tradition of \textit{measurement modeling} \citep{loevinger1957, messick1987, jackman2008}, which has been used in many social science disciplines to build trust in measures of complex social constructs like bias.
Following \citet{messick1987} and \citet{jackman2008}, we say a measure is trustworthy if it is simultaneously \textit{valid} and \textit{reliable}. 
Across disciplines and decades, individual criteria have been defined and refined to designate the key criteria for validity and reliability (see \autoref{tab:measurement-modeling}; we closely follow \citet{jacobs2021}).
For each criteria, we systematically build tests: each test provides incremental evidence for trust, and measures that fare well under all tests accrue considerable evidence to trust them. 

\noindent \textbf{Experimental Details.}
Along with testing our measure, we test measures from prior work, so we reuse the target concepts, social groups, and word lists from prior work (see \autoref{app:reproducibility}).
For target concepts, we follow \citet{garg2018}, drawing upon 104 professions tracked in the US Census \citep{levanon2009}. 
For social groups, we follow \citet{garg2018}, considering either binary gender (female, male) or three-class race/ethnicity (White, Hispanic, Asian). 
For word lists, we follow \citet{bommasani2020}.
In several experiments, we report correlations: Spearman $\rho$ to measure monotonicity, Pearson $R^2$ to measure linearity, and \textbf{bold} to indicate statistic significance for $p \leq 0.05$.

%% file: tables/measurement-modeling.table.tex
\begin{table*}[htp]
\resizebox{\textwidth}{!}{
\vspace{0mm}
\begin{tabular}{c|l|l}
\toprule
\multirow{6}{*}{Validity} & \textbf{Face validity} & Measure passes basic sanity checks.\\
& \textbf{Content validity} & Measure faithfully reflects theoretical understanding of the construct.\\
& \textbf{Convergent validity} & Measure correlates with other credible measures of the same construct. \\
& \textbf{Predictive validity} & Measure predicts other credible measures of related constructs. \\
& \textbf{Hypothesis validity} & Measure enables scientific inquiry related to the construct.\\ 
& \textbf{Consequential validity} & Measure's eventual usage amounts to desirable social impact.\\
\midrule
\multirow{2}{*}{Relability} & \textbf{Inter-annotator agreement} & Measurements are stable up to difference in annotators. \\ 
& \textbf{Sensitivity} & Measurements are stable up to difference in (hyper)parameters. \\ 
\bottomrule
\end{tabular}}
  \caption{Definitions for the 8 measurement modeling criteria we test for in our  testing protocol.}
\label{tab:measurement-modeling}
\end{table*}

%% file: sections/5-validity.tex
\section{Testing Protocol for Validity}
\label{subsec:validity}
\noindent \textbf{Face validity} requires that the measure passes the "sniff test" \citep{jacobs2021}.
To validate our measures in this aspect, we measure gender bias for strongly gender-stereotyped professions (based on heavily imbalanced labor statistics in the 2000 US Census).
In \autoref{tab:facevalidity}, we quantify associations in English Wikipedia, Word2Vec \citep{mikolov2013} and GloVe \citep{pennington2014} embeddings, and BERT \citep[final layer of BERT-base;][]{devlin2019} contextualized representations applied to English Wikipedia.
To measure bias, we juxtapose these observed associations with reference associations of the uniform distribution (i.e.~professions being equally associated with both the female and male gender).
For all professions, across all settings, the measurements align with prevalent US stereotypes, except for \textit{librarian} in settings involving English Wikipedia.
While currently female-stereotyped in the US, the male-leaning stereotype in relation to Wikipedia is justifiable, as most librarians discussed in Wikipedia refer to high-ranking posts (e.g.~Librarian of Congress, University Librarians) historically filled mostly by men. 
\input{tables/facevalidity.table}

\noindent \textbf{Content validity} requires that the measure reflects theoretical understanding of the underlying construct; the measure's structure should match the construct's structure \citep[\textit{structural fidelity};][]{loevinger1957}.
Given the  clear and high-fidelity correspondence between our social bias definition (\autoref{defn:relative-social-bias}), derived from stated principles in \autoref{sec:principles}, and our framework \framework, we argue our measures demonstrate strong content validity. 

\noindent \textbf{Convergent validity} requires that the proposed measure patterns similarly to other measures of the same construct \citep{campbell1959}.
As \citet{jackman2008} writes, convergence is only valuable if prior measures are (known to be) credible. 
Since no prior measure has been subject to rigorous and extensive testing, and measures have been shown to produce drastically different outcomes \citep{bommasani2020}, we find this criterion does not apply directly.

Instead, we reinterpret convergent validity in the context of our measures for bias in text. 
Specifically, while there is no known credible bias measure for text, we introduce two bias measures for text, based on either human or automated judgments (i.e.~cooccurrence). 
Since we consider human judgments to be ideal, we report the correlation between the human and automated measures in \autoref{tab:convergentvalidity}.
Specifically, we measure binary gender bias for  eight professions (those in \autoref{tab:facevalidity}) in the context of English Wikipedia with the uniform distribution as the reference.
Additionally, because we hypothesized human annotators may make more holistic judgments based on context, whereas automated cooccurrence statistics may be more brittle, we also consider the impact of context length in \autoref{tab:convergentvalidity}.
We observe strong correlations for all context lengths and report subsequent results using 3-sentence contexts, given that the strongest correlations occur in this setting. 
\input{tables/convergentvalidity.table} 

\noindent \textbf{Predictive validity} considers whether the measure is predictive of measures of related constructs.
Since social bias is attributed to domain-general cognitive processes \citep{tajfel1969}, we expect that human biases will manifest similarly across different human behaviors. 
Consequently, biases in linguistic performance (e.g.~writing) should predict biases in decision-making (e.g.~employment).\footnote{We clarify that our analyses are strictly correlation-based and not causal. 
Further, perfect predictability is not expected, since it is reasonable that biases in text and hiring are not perfectly correlated, but we do expect significant correlation.}

As a first experiment (\textbf{Diachronic}), we report the correlation between (i) the average bias for 104 professions considered in the US census in Word2Vec embeddings trained on corpora from each decade of 1900--2000 \citep{hamilton2016} and (ii) bias in US labor statistics for the corresponding decades \citep{levanon2009}.
As a second experiment (\textbf{Contemporary}), we report the correlation between bias measurements for each of 104 professions
based on (i) our measurements for contemporary Word2Vec embeddings and (ii) the 2010 Census labor statistics.
In \autoref{tab:predictivevalidity}, we see our measurements consistently track biases in hiring practices, with statisitically significant correlations, whereas several other measures \citep[e.g.][]{bolukbasi2016, ethayarajh2019} do not.
We believe these results hinge on implementation differences for bias in embeddings: the highly-correlated measures all average embeddings/bias scores, whereas the weakly-correlated measures all use PCA.
Strikingly, the measure of \citet{manzini2019}, which is the only other measure that generalizes to the multi-class setting, is strongly \textit{anti-correlated} with diachronic/historical trends in employment for race and gender. 
We return to this measure in \autoref{sec:related-work}, showing it lacks content validity (i.e.~is structurally unfaithful to to the construct of bias) that likely explains its poor predictive validity.
\input{tables/predictivevalidity.table}
\noindent \textbf{Hypothesis validity} requires the measure be useful for addressing scientific hypotheses. 
We study \textit{bias amplification} and \textit{bias mitigation}, since both are central to the social impact of NLP.
\input{tables/overamplification.table}

\noindent \textit{Bias Amplification.} For bias amplification, we test whether training language models, as well as generating text using language models, increases bias.
There is a prevalent hypothesis that model training generally increases bias, with some evidence of this in particular settings in NLP \citep{zhao2017, jia2020}. 
To test this hypothesis, we consider GPT-2 Medium \citep{radford2019}, a publicly available language model, and contrast the associations in GPT-2's training data $\mathcal{L}_1$, GPT-2's contextualized representations $\mathcal{L}_2$ (taken from the final layer), and machine-generated text $\mathcal{L}_3$ sampled from GPT-2.  
Due to the stochasticity involved in sampling, we use a large sample of 250000 unconditional generations from GPT-2.\footnote{\url{https://github.com/openai/gpt-2-output-dataset}}
This experiment highlights the benefits of relative bias measurement, i.e.~requiring an explicit reference, as the effects of processes (training, sampling) can be directly measured.
We use our automated method to measure associations in the (human-authored) training corpus $\mathcal{L}_1$ and machine-generated text corpus $\mathcal{L}_3$; we use probing to measure associations in the contextualized representations $\mathcal{L}_2$ when applied to $\mathcal{L}_1$.
\autoref{tab:overamplification} shows that representation learning in GPT-2 amplifies gender biases relative to the training data, but that much of this bias does not manifest during generation.
Surprisingly, machine-generated text from GPT-2 is measured to be marginally less gender biased than the data used to train GPT-2, which complicates the prevalent hypothesis that learning reliably amplifies the bias in the training data.

\noindent \textit{Bias Mitigation.}~Most "debiasing" methods target word embeddings, generally by directly optimizing a bias measure to provably guarantee bias reduction under that measure \citep[e.g.][]{bolukbasi2016, zhao2018, manzini2019}.
This brings to mind Strathern's law: \textit{``When a measure becomes a target, it ceases to be a good measure''} \citep{strathern1997, goodhart1984}. 
Since we have provided significant evidence to trust our measure, we report in \autoref{tab:debiasing} how mitigation methods change bias according to both our measure and the measure considered in prior work.
While every method reduces bias for the targeted measure, we find that for seven of the eight methods, bias is not reduced and is instead amplified according to ours.
Our findings significantly strengthen existing findings that "debiasing" methods are quite limited  \citep[e.g.][]{gonen2019}: how bias is measured can change, and in many cases invert, judgments about the efficacy of bias mitigation methods.
\input{tables/debiasing.table}

\noindent \textbf{Consequential validity} emphasizes the eventual usage and impact of the measure \citep{messick1988}.
While most of these consequences will be determined in the future, we have proactively implemented our bias measures as the default metrics in the HELM benchmark \citep{liang2022helm} to help accelerate this adoption process.
Already our measures have been used to evaluate 30+ prominent language models to understand model biases across a range of different uses \citep{liang2022helm}.
We will monitor our measures to revisit this question once further evidence accrues on their impact. 

%% file: tables/facevalidity.table.tex
\begin{table}[htp]
\centering
\resizebox{0.47\textwidth}{!}{
\vspace{0mm}
  \begin{tabular}{lcccccc}
    \toprule
& \multicolumn{2}{c}{\textsc{text}} & \multicolumn{2}{c}{\textsc{emb}} & \multicolumn{2}{c}{\textsc{cr}} \\
& Human & Aut. & \textsc{w2v} & \textsc{GloVe} & Red. & Probe \\
\midrule
carpenter & -0.5 & -0.368 & -0.128 & -0.05 & -0.02 & -0.384 \\ 
dancer & 0.167 & 0.039 & 0.078 & 0.086 & 0.035 & 0.09 \\ 
librarian & -0.105 & -0.275 & 0.177 & 0.124 & -0.003 & -0.333 \\ 
nurse & 0.373 & 0.097 & 0.119 & 0.114 & 0.066 & 0.111 \\ 
pilot & -0.417 & -0.265 & -0.099 & -0.072 & -0.022 & -0.33 \\ 
soldier & -0.473 & -0.358 & -0.041 & -0.065 & -0.025 & -0.389 \\ 
\midrule
businessman & -0.5 & -0.341 & -0.173 & -0.145 & -0.056 & -0.232 \\ 
businesswoman & 0.5 & 0.453 & 0.174 & 0.385 & 0.058 & 0.5 \\ 
\bottomrule
\end{tabular}}
\caption{
\textbf{Face validity experiment.} 
Female-directed gender bias for gender-stereotyped professions (\textbf{top}) and explicitly gendered professions (\textbf{bottom}) aligns with prevalent US stereotypes.}
  \label{tab:facevalidity}
\end{table}

%% file: tables/convergentvalidity.table.tex
\begin{table}[t]
\centering
\footnotesize
\vspace{0mm}
  \begin{tabular}{ccc}
    \toprule
Context Length (sent.) & $\rho$ & $R^2$ \\
\midrule
1 & \textbf{0.731} & \textbf{0.848} \\
2 & \textbf{0.814} & \textbf{0.858} \\
3 & \textbf{0.898} & \textbf{0.879} \\
4 & \textbf{0.898} & \textbf{0.840} \\
5 & \textbf{0.898} & \textbf{0.829} \\
\bottomrule
\end{tabular}
  \caption{\textbf{Convergent validity experiment.}  High correlations between human and automated text bias measures for all context lengths.}
  \label{tab:convergentvalidity}
\end{table}
%
%
%

%% file: tables/predictivevalidity.table.tex
\begin{table}[htp]
\centering
\resizebox{0.4\textwidth}{!}{
\vspace{0mm}
\begin{tabular}{lcc|cc}
    \toprule
& \multicolumn{2}{c|}{Diachronic} & \multicolumn{2}{c}{Contemporary} \\
& {Gender} & {Race} & {Gender} & {Race} \\
\midrule
\citet{bolukbasi2016} & 0.261 & N/A & 0.047 & N/A \\
\citet{caliskan2017} & \textbf{0.709} & N/A & \textbf{0.505} & N/A \\
\citet[][cosine]{garg2018} & \textbf{0.758} & N/A & \textbf{0.633} & N/A  \\
\citet[][euclidean]{garg2018} & 0.127 & N/A & \textbf{0.553} & N/A \\
\citet{manzini2019} & \textbf{-0.648} & \textbf{-0.903} & 0.193 & \textbf{-0.396} \\
\citet{ethayarajh2019} & 0.261 & N/A & 0.065 & N/A \\
Our Measure & \textbf{0.83} & \textbf{0.842} & \textbf{0.42} & \textbf{0.369} \\
\bottomrule 
\end{tabular}}
  \caption{
  \textbf{Predictive validity experiments.} Our measures demonstrate high Spearman correlation with \textbf{diachronic} changes in labor statistics, as well as \textbf{contemporary} labor statistics, whereas some other measures do not.  
  }
  \label{tab:predictivevalidity}
\end{table}

%% file: tables/overamplification.table.tex
\begin{table}[htp]
\centering
\resizebox{0.4\textwidth}{!}{
\vspace{0mm}
  \begin{tabular}{lccc}
    \toprule
& $\mathcal{L}_2;\mathcal{L}_1$ & $\mathcal{L}_3;\mathcal{L}_2$ & $\mathcal{L}_3;\mathcal{L}_1$ \\
\midrule
carpenter & 0.176 & -0.166 & 0.010 \\ 
dancer & 0.063 & -0.143 & -0.080 \\ 
librarian & 0.127 & -0.146 & -0.019 \\ 
nurse & 0.017 & -0.136 & -0.119 \\ 
pilot & -0.063 & 0.066 & 0.003 \\ 
soldier & -0.041 & 0.089 & 0.049 \\ 
businessman & -0.021 & 0.066 & 0.044 \\ 
businesswoman & 0.118 & -0.086 & 0.032 \\ 
\bottomrule
\end{tabular}}
\caption{\textbf{Hypothesis validity (amplification) experiment.}
GPT-2's learned representations $\mathcal{L}_2$ amplify bias relative to training data $\mathcal{L}_1$ but much of this bias does not persist to (unconditional) machine-generated samples $\mathcal{L}_3$.
}
\label{tab:overamplification}
\end{table}

%% file: tables/debiasing.table.tex
\begin{table}[htp]
\centering
\resizebox{0.45\textwidth}{!}{
\vspace{0mm}
  \begin{tabular}{crccc|cc}
    \toprule
& & & \multicolumn{2}{c|}{Targeted metric} & \multicolumn{2}{c}{Our metric} \\ 
Emb. & Method & Groups & Original & Debiased &  Original & Debiased \\
\midrule
\textsc{w2v}   & Hard (B) & \textit{gender}   & 0.050 & \best{0.041}  & 0.011 & \best{0.004} \\
\textsc{GloVe} & GN (Z)   & \textit{gender}   & 0.191 & \best{0.083}  & 0.009 & \worst{0.016} \\
\textsc{w2v}   & Soft (M) & \textit{gender}   & 0.330 & \best{0.197}  & 0.008 & \worst{0.012} \\
\textsc{w2v}   & Hard (M) & \textit{gender}   & 0.330 & \best{0.281}  & 0.008 & \worst{0.024} \\
\textsc{w2v}   & Soft (M) & \textit{race}     & 0.026 & \worst{-0.055} & 0.018 & \worst{0.018} \\
\textsc{w2v}   & Hard (M) & \textit{race}     & 0.026 & \best{0.005}  & 0.018 & \worst{0.023} \\
\textsc{w2v}   & Soft (M) & \textit{religion} & 0.253 & \best{0.126}  & 0.023 & \worst{0.024} \\
\textsc{w2v}   & Hard (M) & \textit{religion} & 0.253 & \best{0.217}  & 0.023 & \worst{0.074} \\

\bottomrule
\end{tabular}}
\caption{\textbf{Hypothesis validity (debiasing) experiment.} 
Debiasing methods generally reduce bias (green) for the targeted metric, but generally increase bias (red) for our metric. 
B indicates \citet{bolukbasi2016}, Z indicates \citet{zhao2018}, M indicates \citet{manzini2019}; Hard/Soft/GN refer to specific debiasing methods.}  
\label{tab:debiasing}
\end{table}

%% file: sections/6-reliability.tex
\section{Testing Protocol for Reliability}
\label{sec:reliability}
\noindent \textbf{Inter-annotator agreement} is required for measures to be reliable, though it generally refers to measures that involve human judgments \citep{jackman2008}. 
While the majority of our measures are fully automated, we do introduce a method to measure associations in text based on human judgments.
To estimate the inter-annotator agreement, we recruit 5 NLP researchers (unaffiliated with the project) to annotate 40 contexts for binary gender with the targets being the eight professions used throughout this work.
We report a very high inter-annotator agreement of Fleiss' $\kappa = 0.79$ \citep{landis1977} for this task. 

\noindent \textbf{Sensitivity} is not a standard criteria in measurement modeling, to our knowledge, but since our measures involve several inputs and hyperparameters, we study how sensitive each measure is to perturbations of each of these.
In particular, several works \citep{ethayarajh2019, nissim2020, antoniak2021} shows prior bias measures are highly sensitive to word list perturbations. 
However, in \autoref{tab:sensitivity}, we find that all of our measures are quite stable to variations in the word lists, normalization function, and distance function.
\input{tables/sensitivity.table}

%% file: tables/sensitivity.table.tex
\begin{table}[htp]
\centering
\resizebox{0.45\textwidth}{!}{
\vspace{0mm}
  \begin{tabular}{cccccc}
    \toprule
Setting & $\mathcal{L}$ & \multicolumn{2}{|c|}{Word Lists} & \multicolumn{1}{c|}{$D$} & \multicolumn{1}{c}{\normalize} \\
& & \multicolumn{1}{|c}{$|W(G)| = 3$} & $|W(G)| = 5$  & \multicolumn{1}{|c|}{$\ell_2$} & $\texttt{Softmax}$   \\
\midrule
\textsc{text-aut.} & Wiki. & (\textbf{0.85}, \textbf{0.89}) & (\textbf{0.88}, \textbf{0.94}) & (\textbf{1.00}, \textbf{1.00}) & (\textbf{0.84}, \textbf{0.82})\\
\midrule 
\textsc{emb.} & W2V & (\textbf{0.92}, \textbf{0.94}) & (\textbf{0.90}, \textbf{0.93}) & (\textbf{1.00}, \textbf{1.00}) & (\textbf{0.90}, \textbf{0.95}) \\
\textsc{emb.} & GloVe & (\textbf{0.88}, \textbf{0.94}) & (\textbf{0.90}, \textbf{0.95}) & (\textbf{1.00}, \textbf{1.00}) & (\textbf{0.90}, \textbf{0.90}) \\
\midrule 
\textsc{cr-red.} & BERT & (\textbf{0.90}, \textbf{0.98}) & (\textbf{1.00}, \textbf{0.99}) &  (\textbf{1.00}, \textbf{1.00}) &  (\textbf{1.00}, \textbf{1.00}) \\
\textsc{cr-probe} & BERT & N/A & N/A & (\textbf{1.00}, \textbf{1.00}) & (\textbf{0.79}, \textbf{0.71})  \\
\bottomrule
\end{tabular}}
  \caption{\textbf{Sensitivity experiment.}
Perturbing any single parameter/input yields stable results for our measures, when compared to the default parameters, based on $(\rho, R^2)$ correlations. 
For word lists, we subsample lists to the specified size, similar to \citet{ethayarajh2019}.}
\label{tab:sensitivity}
\end{table}

%% file: sections/7-related-work.tex
\section{Related Work}
\label{sec:related-work}

\noindent \textbf{Text.}
In the social sciences, work across many disciplines has qualitatively characterized social bias in specific corpora of interest \citep[\eg][]{blumberg2007, atir2018}.
While several quantitative measures have recently been proposed \citep{rudinger2017, bordia2019, field2020, falenska2021, sun2021, mitchell2022}, to our knowledge, these methods have neither been significantly adopted to facilitate social science research nor to measure bias in NLP datasets.
We find this surprising given especially how large text corpora have been instrumental to the rise of language models in the field \citep[][\textit{inter alia}]{peters2018, devlin2019, brown2020}, alongside growing broader interest in dataset documentation and governance \citep{caswell2021, bandy2021, dodge2021, bender2018, gebru2021, jernite2022}.
For this reason, we apply our measures to bias measurement on both sides of language modeling: the initial human-authored training corpora as well as the final machine-generated samples, and our measures have been similarly applied in the HELM benchmark for many language models and use cases \citep{liang2022helm}. 
Mechanically, our bias measures for text, as well as other bias measures for text \citep[\eg][]{bordia2019}, bear strong resemblance to the estimates of mutual information introduced by \citet{church1989}. 

\noindent \textbf{Representations.}
\citet{bolukbasi2016} initiated the study of bias measurement for word embeddings, with a growing collection of such measures \citep[e.g.][]{bolukbasi2016, caliskan2017, garg2018, ethayarajh2019, manzini2019, du2019, kumar2020}.
More recently, these measures have been adapted to measure bias in contextualized representations, generally by reducing measurement to the word embedding setting \citep{bommasani2020}, either by specifying a singular canonical context \citep{may2019, tan2019, ross2021} or averaging representations across many contexts \citep{bommasani2020, guo2020, steed2021}.
In comparison to prior measures for representations, we delineate the following differences.
First, our measures are the only existing measures that are directly constructed under a unified framework for text and representation bias measurement.
As we show in \autoref{tab:overamplification}, this enable us to study the effects of training (a transformation from text to representations) and generation (a transformation from representations to text). 
Second, all of our measures permit multiclass bias measurement, which is necessary given the underlying social categories are generally non-binary.
To our knowledge, the measure of \citet{manzini2019} (and measures that directly extend it) is the only prior measure that also extends to the multiclass setting.

Given this, we further examined this measure to understand the difference between it and our measures.
Empirically, in \autoref{tab:predictivevalidity}, we found our measure was highly correlated with both diachronic and contemporary trends in employment, whereas the measure of \citet{manzini2019} was either uncorrelated or anti-correlated, indicating it lacks predictive validity.
Further, in \autoref{tab:debiasing}, we found that mitigation methods that successfully optimize for the metric of \citet{manzini2019} always increase bias under our method, independent of the specific optimization method (hard or soft) and the groups considered (i.e.~gender, race, religion). 
Tracing this to the mathematical definition, we find the measure of \citet{manzini2019} lacks content validity (which likely explain the above empirical findings). 
As a minimal example, consider that the binary gender bias according to \citet{manzini2019}'s measure for the concept \textit{scientist}, using the word lists $\{\textit{man}\} \text{ and} \{\textit{woman}\}$, is proportional to:
\begin{align*}
\cos(\mathbf{scientist}, \mathbf{man}) + \cos(\mathbf{scientist}, \mathbf{woman})
\end{align*}
This fails to meet the criteria of content validity and structural fidelity, as it is not faithful to the underlying construct of social bias: social bias is proportional to (as codified in all other measures) the difference in the associations, not the sum. 

\noindent \textbf{Other settings.}
In addition to measuring bias in text and language representations, several recent works investigate biases in language models via the probabilities they assign to specific words or sequences \citep{kurita2019, nangia2020, nadeem2021}.
Since language modeling is currently the premier means for representation learning \citep{devlin2019, bommasani2021}, there is a natural question regarding the relationship between measuring biases of a pretrained language model and of representations induced by a pretrained language model.\footnote{This mirrors the distinction between behavioral and representational methods in interpretability \citep{belinkov2021}.}
In our work, since we are motivated by the potential downstream harms of language technologies, we elect to measure biases in representations as 1) it is the representations that are used downstream and 2) some biases may not manifest in sequence probabilities, but are latently present in the representations, and therefore may still manifest in downstream settings.
To be more explicit, if some biases in the representations remain "dormant" and do not appear during generation (which is precisely what we saw in our experiments with GPT-2 in \autoref{tab:overamplification}), they will be invisible in these behavioral evaluations of language models. 
Nonetheless, these biases could observably affect model behavior once the language model is fine-tuned for downstream tasks, which is likely where the most concerning harms arise. 

Further downstream, fairness evaluations exist for specific tasks such as machine translation \citep{stanovsky2019, escude2019, prates2019}, text generation \citep{sheng2019, gehman2020, dhamala2021, lucy2021}, coreference resolution \citep{rudinger2018, zhao2018b, cao2020}, sentiment analysis \citep{kiritchenko2018}, relation extraction \citep{gaut2020}, and question answering \citep{parrish2021}.\footnote{See \citet{czarnowska2021} for a summary.}
Given the existing paradigm of upstream pretraining and downstream adaptation/fine-tuning, future work should investigate the predictive validity of upstream bias measures at predicting downstream bias measures \citep{goldfarb2021, jin2021}. 

%% file: sections/8-measure-modeling.tex
\section{Discussion of Measurement Modeling}
\label{sec:measurement-modeling-future}
In \autoref{sec:verification}, we stress test our measures using measurement modeling, an interdisciplinary theory with a long history \citep{loevinger1957, messick1987, jackman2008}.
Our work joins a growing collection of recent works that embrace measurement modeling in computational and AI contexts \citep{jacobs2021, milli2021, blodgett2021b}.
For social bias in NLP, recent works use measurement modeling to identify failures in the validity \citep{blodgett2021} and reliability \citep{zhang2020, du2021} of existing bias measures.
In contrast, our work is the first to argue for the trustworthiness of social bias measures based on testing via measurement modeling.
With that said, we emphasize that this does not unequivocally cement the trustworthiness of our measures, especially in contexts they have not been tested in: we have shown our measures pass the tests we introduce, but there certainly may be (and likely are) others that would demonstrate their weaknesses. 

Beyond social bias, we believe measurement modeling can be a powerful general-purpose method in NLP in contexts where measurement/evaluation may be hard (but trust in evaluation is critical).
To briefly demonstrate this, we enumerate several instances where existing work that studies evaluation in a particular context can be reinterpreted as referring to one (or more) of the criteria in measurement modeling.
Critically, none of these works leverage either the specific language, or broader theory, of measurement modeling, but they can all be unified under this lens.
In natural language generation evaluation, numerous works \citep[e.g.][]{zhang2020b, sellam2020, hessel2021,  pillutla2021} argue for their metrics to be used in place of existing metrics like BLEU \citep{papineni2002}, because they more faithfully capture the semantics of language compared to the brittle overlap-based BLEU, and/or they are more correlated with human judgments.
In essence, these works are arguing for the content validity and/or the convergent validity of their metrics. 
In the analysis of explainability methods, \citet{jacovi2020} argue several  methods improperly conflate the plausibility and faithfulness of evaluations, which can be understood as a failure in the content validity of these methods. 
And, in the evaluation of word embeddings, \citet{chiu2016} and \citet{rogers2018} show intrinsic evaluations (e.g.~word analogy tests, word similarity/relatedness) do not reliably correlate with extrinsic evaluations of downstream outcomes (e.g.~the performance of models built using these embeddings), indicating they lack predictive validity.
\citet{ravichander2021} provide similar results for the intrinsic evaluations of syntactic understanding versus downstream behavior on entailment tasks.

More generally, measurement modeling provides a battle-tested set of well-studied desiderata, which can be used to standardize how we evaluate measures in NLP.
In particular, while the criteria in measurement modeling are unlikely to be truly exhaustive, they do represent a comprehensive taxonomy of what properties are important for a measure to satisfy.
In practice, we imagine this would yield an explicit protocol for accruing trust in a measure/evaluation by subjecting the measure/evaluation to a battery of tests \citep[cf. the software engineering tests of][]{ribeiro2020}.

%% file: sections/9-conclusion.tex
\section{Conclusion}
\label{sec:conclusion}
In this work, we foreground trust in social bias measurement: how do we accrue the evidence necessary to warrant trusting bias measures?
Trustworthy bias measures are integral for making progress on broader goals (\eg~harm reduction through bias mitigation), which are of increasing consequence as the footprint of language technology and NLP grows.
Our work contributes a general measurement framework \framework to measure bias, based on principles in social science, along with a testing protocol based on measurement modeling.
Together, this makes the case for our social bias measures being trustworthy.
However, as \citet{messick1987, messick1988} explains, the task of validating a measure is an ongoing process: from the consequentialist perspective, it will be the use of our measures that determines their value. 

%% file: sections/10-reproducibility.tex
\section{Reproducibility}
\label{sec:reproducibility}
All code is made available at \url{https://github.com/rishibommasani/BiasMeasures} with further details on data/sources in \autoref{app:reproducibility}.
We aim to release further tooling to facilitate adoption of our measures in future work along with documentation of the impact of our measures over time, consistent with the discussion of consequential validity.

%% file: sections/acknowledgements.tex
Thanks to Kawin Ethayarajh, Sidd Karamcheti, Nelson Liu, Claire Cardie, Su Lin Blodgett,  Shyamal Buch, Lisa Li, Tianyi Zhang, Xikun Zhang, Shiori Sagawa, Michael Xie, Ananya Kumar, Deep Ganguli, Tatsu Hashimoto, Dan Ho, and members of p-lambda, Stanford NLP and Cornell NLP for feedback on this work.
RB was supported by an NSF Graduate Research Fellowship, under
grant number DGE-1656518. 
Other funding was provided by a PECASE award to PL.

%% file: appendices/01_reproducibility.tex
\section{Reproducibility}
\label{app:reproducibility}

\subsection{Word Lists}

We use social groups from \citet{garg2018} and word lists from \citet{bommasani2020}, which we explicitly enumerate consistent with the recommendations of \citet{antoniak2021}.

\texttt{Female} word list = \{she', `daughter', `hers', `her', `mother', `woman', `girl', `herself', `female', `sister', `daughters', `mothers', `women', `girls', `femen', `sisters', `aunt', `aunts', `niece', `nieces' $\}$ \\

\texttt{Male} word list = \{`he', `son', `his', `him', `father', `man', `boy', `himself', `male', `brother', `sons', `fathers', `men', `boys', `males', `brothers', `uncle', `uncles', `nephew', `nephews' $\}$ \\

\texttt{Asian} word list = \{`cho', `wong', `tang', `huang', `chu', `chung', `ng', `wu', `liu', `chen', `lin', `yang', `kim', `chang', `shah', `wang', `li', `khan', `singh', `hong'$\}$ \\

\texttt{Hispanic} word list = \{`castillo', `gomez', `soto', `gonzalez', `sanchez', `rivera', `martinez', `torres', `rodriguez', `perez', `lopez', `medina', `diaz', `garcia', `castro', `cruz' $\}$ \\

\texttt{White} word list = \{`harris', `nelson', `robinson', `thompson', `moore', `wright', `anderson', `clark', `jackson', `taylor', `scott', `davis', `allen', `adams', `lewis', `williams', `jones', `wilson', `martin', `johnson' $\}$ \\

\texttt{Professions} word list = \{`accountant', `acquaintance', `actor', `actress', `administrator', `adventurer', `advocate', `aide', `alderman', `ambassador', `analyst', `anthropologist', `archaeologist', `archbishop', `architect', `artist', `artiste', `assassin', `astronaut', `astronomer', `athlete', `attorney', `author', `baker', `ballerina', `ballplayer', `banker', `barber', `baron', `barrister', `bartender', `biologist', `bishop', `bodyguard', `bookkeeper', `boss', `boxer', `broadcaster', `broker', `bureaucrat', `businessman', `businesswoman', `butcher', `cabbie', `cameraman', `campaigner', `captain', `cardiologist', `caretaker', `carpenter', `cartoonist', `cellist', `chancellor', `chaplain', `character', `chef', `chemist', `choreographer', `cinematographer', `citizen', `cleric', `clerk', `coach', `collector', `colonel', `columnist', `comedian', `comic', `commander', `commentator', `commissioner', `composer', `conductor', `confesses', `congressman', `constable', `consultant', `cop', `correspondent', `councilman', `councilor', `counselor', `critic', `crooner', `crusader', `curator', `custodian', `dad', `dancer', `dean', `dentist', `deputy', `dermatologist', `detective', `diplomat', `director', `doctor', `drummer', `economist', `editor', `educator', `electrician', `employee', `entertainer', `entrepreneur', `environmentalist', `envoy', `epidemiologist', `evangelist', `farmer', `filmmaker', `financier', `firebrand', `firefighter', `fireman', `fisherman', `footballer', `foreman', `gangster', `gardener', `geologist', `goalkeeper', `guitarist', `hairdresser', `handyman', `headmaster', `historian', `hitman', `homemaker', `hooker', `housekeeper', `housewife', `illustrator', `industrialist', `infielder', `inspector', `instructor', `inventor', `investigator', `janitor', `jeweler', `journalist', `judge', `jurist', `laborer', `landlord', `lawmaker', `lawyer', `lecturer', `legislator', `librarian', `lieutenant', `lifeguard', `lyricist', `maestro', `magician', `magistrate', `manager', `marksman', `marshal', `mathematician', `mechanic', `mediator', `medic', `midfielder', `minister', `missionary', `mobster', `monk', `musician', `nanny', `narrator', `naturalist', `negotiator', `neurologist', `neurosurgeon', `novelist', `nun', `nurse', `observer', `officer', `organist', `painter', `paralegal', `parishioner', `parliamentarian', `pastor', `pathologist', `patrolman', `pediatrician', `performer', `pharmacist', `philanthropist', `philosopher', `photographer', `photojournalist', `physician', `physicist', `pianist', `planner', `playwright', `plumber', `poet', `policeman', `politician', `pollster', `preacher', `president', `priest', `principal', `prisoner', `professor', `programmer', `promoter', `proprietor', `prosecutor', `protagonist', `protege', `protester', `provost', `psychiatrist', `psychologist', `publicist', `pundit', `rabbi', `radiologist', `ranger', `realtor', `receptionist', `researcher', `restaurateur', `sailor', `saint', `salesman', `saxophonist', `scholar', `scientist', `screenwriter', `sculptor', `secretary', `senator', `sergeant', `servant', `serviceman', `shopkeeper', `singer', `skipper', `socialite', `sociologist', `soldier', `solicitor', `soloist', `sportsman', `sportswriter', `statesman', `steward', `stockbroker', `strategist', `student', `stylist', `substitute', `superintendent', `surgeon', `surveyor', `teacher', `technician', `teenager', `therapist', `trader', `treasurer', `trooper', `trucker', `trumpeter', `tutor', `tycoon', `undersecretary', `understudy', `valedictorian', `violinist', `vocalist', `waiter', `waitress', `warden', `warrior', `welder', `worker', `wrestler', `writer' \}

\subsection{Data sources}

\paragraph{English Wikipedia.}
We use the same subset of English Wikipedia that was used by \citet{bommasani2020}, which was chosen because it filtered for bot-generated content, and was sourced from \url{https://blog.lateral.io/2015/06/the-unknown-perils-of-mining-wikipedia/}. 

\paragraph{Contemporary Word Embeddings.}
We use standard static word embeddings: GloVe embeddings (Wikipedia 2014 + Gigaword 5, 300 dimensional) sourced from \url{https://nlp.stanford.edu/projects/glove/} and Word2Vec embeddings (GoogleNews, 300 dimensional) sourced from \url{https://code.google.com/archive/p/word2vec/}. 

\paragraph{Contextualized Representations.}
We use BERT and GPT-2 representations from the checkpoints made available through HuggingFace Transformers \citep{wolf2020}.

\paragraph{GPT-2 related text.}
We use the sample of both GPT-2's training corpus (actually the identically distributed test) and its (unconditional) samples made available at \url{https://github.com/openai/gpt-2-output-dataset}.

\paragraph{US Census data.}
We use the US Census data of \citet{levanon2009} that was also used by \citet{garg2018}, made available at \url{https://github.com/nikhgarg/EmbeddingDynamicStereotypes/tree/master/data}.

\paragraph{Historic Word Embeddings.}
We use the word embeddings trained on different decades in the 1900s from \citet{hamilton2016} that were also used by \citet{garg2018}, made available at \url{https://nlp.stanford.edu/projects/histwords/}.